\documentclass[11pt,letterpaper]{article}
\usepackage{emnlp2017}
\usepackage{times}
\usepackage{latexsym}
\usepackage{graphicx}
\usepackage{CJK,algorithm,algorithmic,amssymb,amsmath,amsfonts,array,graphics,enumerate,multirow,array,float,subfigure,verbatim,epstopdf,tabularx}
\usepackage{times}
\usepackage{latexsym}
\usepackage{enumitem}
\usepackage{url}
\usepackage{color}
\usepackage{hhline}

\emnlpfinalcopy

\def\emnlppaperid{***}

\newcommand{\vpara}[1]{\vspace{0.05in}\noindent\textbf{#1 }}

\definecolor{orange}{rgb}{0.85,0.3,0}

\title{Improving Slot Filling Performance with Attentive Neural Networks on Dependency Structures}

\author{Lifu Huang \\ Rensselaer Polytechnic Institute \\ huangl7@rpi.edu
        \And
        Avirup Sil \\ IBM T.J. Watson Research Center  \\ avi@us.ibm.com \AND
        Heng Ji \\ Rensselaer Polytechnic Institute \\ jih@rpi.edu
        \And
        Radu Florian \\ IBM T.J. Watson Research Center  \\ raduf@us.ibm.com
        }

\date{}

\begin{document}

\maketitle


\begin{abstract}

Slot Filling (SF) aims to extract the values of certain types of attributes (or slots, such as person:cities\_of\_residence) for a given entity from a large collection of source documents. 
In this paper we propose an effective DNN architecture for SF with the following new strategies: (1). Take a regularized dependency graph instead of a raw sentence as input to DNN, to compress the wide contexts between query and candidate filler; (2). Incorporate two attention mechanisms: local attention learned from query and candidate filler, and global attention learned from external knowledge bases, to guide the model to better select indicative contexts to determine slot type. Experiments show that this framework outperforms state-of-the-art on both relation extraction (16\% absolute F-score gain) and slot filling validation for each individual system (up to 8.5\% absolute F-score gain). 

\end{abstract}
\section{Introduction}
\label{sec:intro}

The goal of Slot Filling (SF) is to extract pre-defined types of attributes or slots (e.g., \textit{per:cities\_of\_residence}) for a given query entity from a large collection of documents. The slot filler (attribute value) can be an entity, time expression or value (e.g., \textit{per:charges}). The TAC-KBP slot filling task~\cite{Ji2011,surdeanu2014overview} defined 41 slot types, including 25 types for person and 16 types for organization.

One critical component of slot filling is relation extraction, namely to classify the relation between a pair of query entity and candidate slot filler into one of the 41 types or none. Most previous studies have treated SF in the same way as within-sentence relation extraction tasks in ACE~\footnote{http://www.itl.nist.gov/iad/mig/tests/ace/} or SemEval~\cite{hendrickx2009semeval}. They created training data based on crowd-sourcing or distant supervision, and then trained a multi-class classifier or multiple binary classifiers for each slot type based on a set of hand-crafted features. 

Although Deep Neural Networks (DNN) such as Convolutional Neural Networks (CNN) and Recurrent Neural Networks (RNN) have achieved state-of-the-art results on within-sentence relation extraction~\cite{Zeng2014,Liu2015,santos2015classifying,Nguyen2015a,Yang2016,wangrelation}, there are limited studies on SF using DNN.~\newcite{adelcis} and~\newcite{adel2016comparing} exploited DNN for SF but did not achieve comparable results as traditional methods. In this paper we aim to answer the following questions: What is the difference between SF and ACE/SemEval relation extraction task? How can we make DNN work for SF?

\begin{figure}[h]
\centering
\includegraphics[width=.40\textwidth]{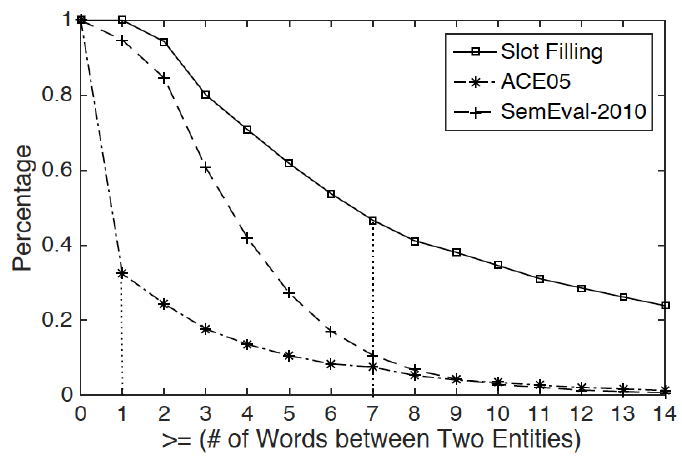}
\caption{Comparison of the Percentage by the \# of Words between two entity mentions in ACE05 and SemEval-2010 Task 8 relations, and between query and slot filler in KBP2013 Slot Filling.}
\label{comparison1}
\end{figure}

We argue that SF is different and more challenging than traditional relation extraction. First, a query and its candidate filler are usually separated by much wider contexts than the entity pairs in traditional relation extraction. As Figure~\ref{comparison1} shows, in ACE data, for 70\% of relations, two mentions are embedded in each other or separated by at most one word. In contrast, in SF, more than 46\% of $\langle$\textit{query, filler}$\rangle$ entity pairs are separated by at least 7 words. For example, in the following sentence:
\begin{enumerate}[label={E}\arabic*.]
\item ``\textbf{Arcandor}$_{query}$ owns a 52-percent stake in Europe's second biggest tourism group Thomas Cook, the Karstadt chain of department stores and iconic shops such as the \textbf{KaDeWe}$_{filler}$ in what used to be the commercial heart of West Berlin.'', 
\end{enumerate}
Here, \textbf{Arcandor} and \textbf{KaDeWe} are far separated and it's difficult to determine the slot type as \textit{org:subsidiaries} based on the raw wide contexts. 

In addition, compared with relations defined in ACE (18 types) and SemEval (9 types), slot types are more fine-grained and heavily rely on indicative contextual words for disambiguation. ~\newcite{yu2015read} and~\newcite{yuunsupervised} demonstrate that many slot types can be specified by contextual \textit{trigger} words. Here, a \textit{trigger} is defined as the word which is related to both the query and candidate filler, and can indicate the type of the target slot. 
Considering E$1$ again, \textit{owns} is a trigger word between \textbf{Arcandor} and \textbf{KaDeWe}, which can indicate the slot type as \textit{org:subsidiaries}. Most previous work manually constructed trigger lists for each slot type. However, for some slot types, the triggers can be implicit and ambiguous.

To address the above challenges, we propose the following new solutions:

\begin{enumerate}[label=$\bullet$]
\item 
To compress wide contexts, we model the connection of query and candidate filler using dependency structures, and feed dependency graph to DNN. To our knowledge, we are the first to directly take dependency graphs as input to CNN. 

\item 
Motivated by the definition of \textit{trigger}, we design two attention mechanisms: a local attention and a global attention using large external knowledge bases (KBs), to better capture implicit clues that indicate slot types. 
\end{enumerate}

\section{Architecture Overview}
\label{sec:architecture}

\begin{figure*}[!htb]
\centering
\includegraphics[width=.99\textwidth]{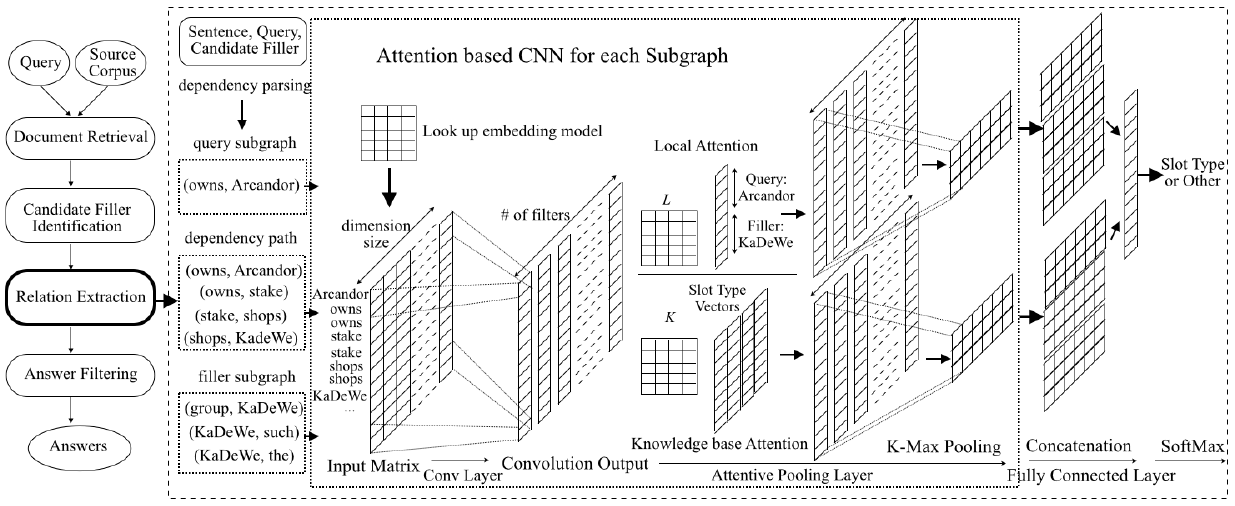}
\vspace{-0.2cm}
\caption{Overview of the Architecture.}
\label{overview}
\vspace{-0.2cm}
\end{figure*}

Figure~\ref{overview} illustrates the pipeline of a SF system. 
Given a query and a source corpus, the system retrieves related documents, identifies  candidate fillers (including entities, time, values, and titles), extracts the relation between query and each candidate filler occurring in the same sentence, and finally determines the filler for each slot. Relation extraction plays a vital role in such a SF pipeline. In this work, we focus on relation extraction component and design a neural architecture. 

Given a query, a candidate filler, and a sentence, we first construct a regularized dependency graph and take all $\langle$\textit{governor, dependent}$\rangle$ word pairs as input to Convolutional Neural Networks (CNN). 

Moreover, We design two attention mechanisms: \textbf{(1) Local Attention}, which utilizes 
the concatenation of \textit{Query} and \textit{Candidate Filler} vectors 
to measure the relatedness of each input bigram (we set filter width as 2) to the specific query and filler. 
\textbf{(2) Global attention}: We use pre-learned slot type representations to 
measure the relatedness of each input bigram with each slot type via a transformation matrix. 
These two attention mechanisms will guide the pooling step to select the information which is related to query and filler and can indicate slot type. 



\section{Regularized Dependency Graph based CNN}
\subsection{Regularized Dependency Graph}
\label{sec:depreg}


Dependency parsing based features, especially the shortest dependency path between two entities, have been proved to be effective to extract the most important information for identifying the relation between two entities~\cite{bunescu2005shortest,zhao2005extracting,guodong2005exploring,jiang2007systematic}. Several recent studies also explored transforming a  dependency path into a sequence and applied Neural Networks to the sequence for relation classification~\cite{Liu2015,caibidirectional,xu2015classifying}.  



However, for SF, the shortest dependency path between query and candidate filler is not always sufficient to infer the slot type due to two reasons. First, the most indicative words may not be included in the path. For example, in the following sentence: 


\begin{enumerate}[label={E}\arabic*.]\setcounter{enumi}{1}
\item Survivors include two \textit{sons} and \textit{daughters}-in-law, \textbf{Troy}$_{filler}$ and Phyllis Perry, \textbf{Kenny}$_{query}$ and Donna Perry, all of Bluff City.
\end{enumerate}
the shortest dependency path between \textbf{Kenny} and \textbf{Troy} is:
``Troy $\leftarrow^{conj}$ Perry $\leftarrow^{conj}$ Kenny'', which does not include the most indicative words: \textit{sons} and \textit{daughters} for their \textit{per:siblings} relation. 
In addition, the relation between query and candidate filler is also highly related to their entity types, especially for disambiguating slot types such as \textit{per:country\_of\_birth} \textit{per:state\_of\_birth} and \textit{per:city\_of\_birth}. Entity types can be inferred by enriching query and filler related contexts. For example, in the following sentence: 


\begin{enumerate}[label={E}\arabic*.]\setcounter{enumi}{2}
\vspace{-0.1cm}
\item \textbf{Merkel}$_{query}$ died in the southern German city of \textbf{Passau}$_{filler}$ in 1967.
\end{enumerate}
we can determine the slot type as  \textit{city} related by incorporating rich contexts (e.g., ``\textit{city}''). 


To tackle these problems, we propose to regularize the dependency graph, incorporating the shortest dependency path between query and candidate filler, as well as their rich contextual words. 





\begin{figure}[h]
\centering
\includegraphics[width=.35\textwidth]{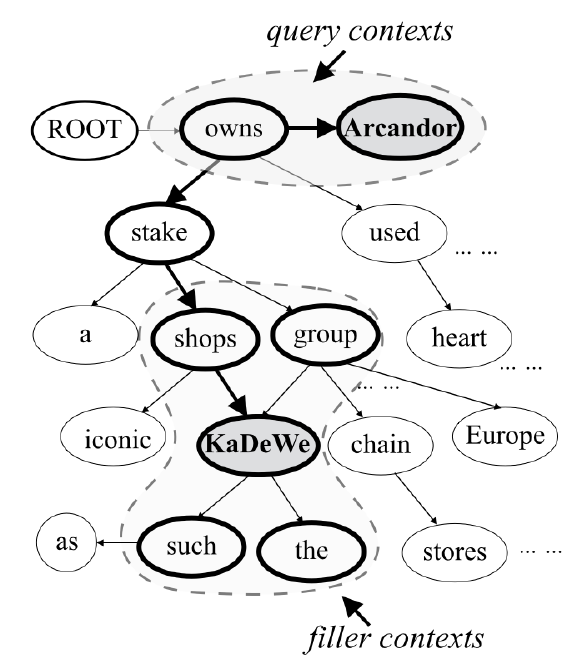}
\vspace{-0.1cm}
\caption{Regularized Dependency Graph for Query \textit{Arcandor} and Filler \textit{KaDeWe} in E1.}
\vspace{-0.1cm}
\label{dependencyParsing}
\end{figure}


Given a sentence $s$ including a query $q$ and candidate filler $f$, we first apply the Stanford Dependency Parser~\cite{Manning2014} to generate all dependent word pairs: $\langle$\textit{governor, dependent}$\rangle$, then discover the shortest dependency path between query and candidate filler based on Breadth-First-Search (BFS) algorithm. The regularized dependency graph includes words on the shortest dependency path, as well as words which can be connected to query and filler within $n$ hops. In our  experiments, we set $n=1$. Figure~\ref{dependencyParsing} shows the dependency parsing output for E1 mentioned in Section~\ref{sec:intro}, and the regularized dependency graph with the bold circled nodes. We can see that, the most indicative trigger \textit{owns} can be found in both the shortest dependency path of \textbf{Arcandor} and \textbf{KaDeWe}, and the context words of \textbf{Arcandor}. In addition, the context words, such as \textit{shops}, can also infer the type of candidate filler \textbf{KaDeWe} as an Organization.



\subsection{Graph based CNN}
\label{sec:basiccnn}




Previous work~\cite{adel2016comparing} split an input sentence into three parts based on the positions of the query and candidate filler and generate a feature vector for each part using a shared CNN. 
To compress the wide contexts, instead of taking the raw sentence directly as input, we split the regularized dependency graph into three parts: query related subgraph, candidate filler related subgraph, and the dependency path between query and filler. Each subgraph will be taken as input to a CNN, as illustrated in 
Figure~\ref{overview}. 
We now describe the details of each part as follows.




\vpara{Input layer:}
Each subgraph or path $G$ in the regularized dependency graph is represented as a set of dependent word pairs $G = \{\langle g_1, d_1\rangle, \langle g_2, d_2\rangle, ... \langle g_n, d_n\rangle\}$. Here, $g_i, d_i$ denote the \textit{governor} and \textit{dependent} respectively. 
Each word is represented as a $d$-dimensional pre-trained vector. For the word which does not exist in the pre-trained embedding model, we assign a random vector for it. Each word pair $\langle g_i, d_i\rangle$ is converted to a $\mathbb{R}^{2\times d}$ matrix. We concatenate the matrices of all word pairs and get the input matrix $M\in \mathbb{R}^{2n\times d}$. 


\vpara{Convolution layer:}For each subgraph, $M\in \mathbb{R}^{2n\times d}$ is the input of the convolution layer, 
which is a list of linear layers with parameters shared by filtering windows with various size. We set the stride as 2 to obtain all word pairs from the input matrix $M$. For each word pair $p_{in}=\langle v_{g_i}, v_{d_i}\rangle$, we compute the output vector $p_{out}$ of a convolution layer as: 
\begin{displaymath}
p_{out} = tanh(W\cdot p_{in} + b)
\end{displaymath}
where $p_{in}$ is the concatenation of vectors for the words $v_{g_i}$ and $v_{d_i}$, $W$ denotes the convolution weights, and $b$ is the bias. In our work all three convolution layers share the same $W$ and $b$.

\vpara{K-Max Pooling Layer:} 
we follow~\newcite{adel2016comparing} and use K-max pooling to select K values for each convolution layer. Later we will incorporate attention mechanisms into K-max pooling. 

\vpara{Fully Connected Layer:}
After getting the high-level features based on the (attentive) pooling layer for each input subgraph, we flatten and concatenate these three outputs as input to a fully connected layer. This layer connects each input to every single neuron it contains, and learns non-linear combinations based on the whole input. 

\vpara{Output Layer:} 
It takes the output of the fully connected layer as input to a softmax regression function to predict the type. We use negative log-likelihood as loss function to train the parameters. 


\section{Attention Strategies for SF}
\subsection{Local Attention}
\label{sec:local}


The basic idea of attention mechanism is to assign a weight to each position of a lower layer when computing the representations for an upper layer, so that the model can be attentive to specific regions~\cite{bahdanau2014neural}. 
In SF, the indicative words are the most meaningful information that the model should pay attention to.~\newcite{wangrelation} applied attention from the entities directly to determine the most influential parts in the input sentence. Following the same intuition, we apply the attention from the query and candidate filler to the convolution output instead of the input, to avoid information vanishing during convolution process~\cite{yin2015abcnn}. 



\begin{figure}[!htp]
\centering
\includegraphics[width=.41\textwidth]{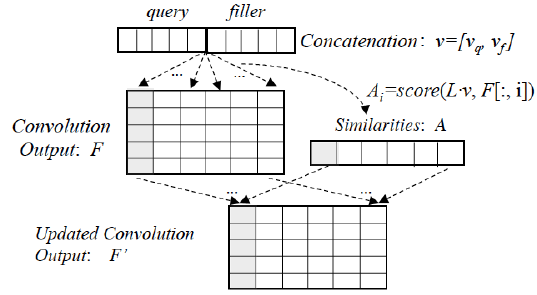}
\vspace{-0.2cm}
\caption{Local Attention.}
\label{localattention}
\vspace{-0.3cm}
\end{figure}

Figure~\ref{localattention} illustrates our approach to incorporate local attention. 
We first concatenate the vector of query $q$ and candidate filler $f$ using pre-trained embeddings $v = [v_q, v_f], \in \mathbb{R}^{2d}$. For $q$ or $f$ that includes multiple words, we average the vectors of all individual words. 
For each convolution output $F$, which is a feature map $\in \mathbb{R}^{\widehat{K}\times N}$, where $N$ is the number of word pairs from the input, and $\widehat{K}$ is the number of filters
, we define the attention similarity matrix $A\in R^{N\times 1}$ as:
\begin{displaymath}
A_{i} = cosine(L\cdot v, F[:,i])
\end{displaymath}
where $L\in \mathbb{R}^{\widehat{K}\times 2d}$ is the transformation matrix between the concatenated vector $v$ and convolution output
. $F[:,i]$ denotes the vector of column $i$ in $F$. 
Then we use the attention matrix $A$ to update each column of the feature map $F$, and generate an updated attention feature map $F^{'}$ as follows:
\begin{displaymath}
F^{'}[:,i] = F[:, i]\cdot A[i]
\end{displaymath}

\subsection{Global Attention}
\label{sec:kb}





Considering E1 in Section~\ref{sec:intro} again, the most discriminating word \textit{owns} 
is not only related to the query and filler, but more specific to the type \textit{org:subsidiaries}. Local attention aims to identify the query and filler related contexts. In order to detect type-indicative parts, we design global attention, using pre-learned slot type representations.

\newcite{wangrelation} explored relation type attention with automatically learned type vectors from training data. However, in most cases, the training data is not balanced and some relation types cannot be assigned high-quality vectors with limited data. Thus, we designed two methods to generate pre-learned slot type representations. 

First, we compose pre-trained lexical word embeddings of each slot type name to directly generate type representations. For example, for the type \textit{per:date\_of\_birth}, we average the vectors of all single tokens (\emph{person}, \emph{birth}, \emph{date}) within the type name as its representation.

Another new method is 
to take advantage of the large size of facts from external knowledge base (KB) to represent slot types. We use DBPedia as the target KB and manually map KB relations to slot types.  
For example, \textit{per:alternate\_names} can be mapped to \textit{alternativeNames}, \textit{birthName} and \textit{nickName} 
 in DBPedia. Thus for each slot type, we collect many triples: $\langle$\textit{query, slot, filler}$\rangle$ and use TransE~\cite{bordes2013translating}, which models slot types as translations operating on the embeddings of query and filler, to derive a representation for each slot type. Compared with the first lexical based slot type representation induction approach, TransE jointly learns entity and relation representations and can better capture the correlation and differentiation among various slot types. Later, we will show the impact of these two types of slot type representations in Section~\ref{sec:comparison1}.



Next we use the pre-learned slot type representations to guide the pooling process. Formally, let $R\in \mathbb{R}^{d\times r}$ be the matrix of all slot type vectors, where $d$ is the vector dimension size and $r$ is the number of slot types. Let $F\in{\mathbb{R}^{\widehat{K}\times N}}$ be a convolution output, which is the same as Section~\ref{sec:local}. We define the attention weight matrix $S$ as:
\begin{displaymath}
S_{i, j} = cosine(F[:, i], W\cdot R[:, j])
\end{displaymath}
where $W\in \mathbb{R}^{\widehat{K}\times d}$ is the transformation matrix for pre-learned slot type representations and convolution output. Given the weight matrix $S$, we generate the attention feature map $F^{''}$ as follows:
\begin{displaymath}
F^{''}[:, i] = F[:, i]\cdot \max_{j}\{S[i, j]\}
\vspace{-0.2cm}
\end{displaymath}
where $S[i, :]$ denotes the similarity scores between column $i$ in $F$ with all slot type vectors, and $\max\{S[i, :]\}$ is the $max$ value among all similarity scores for column $i$ in $F$. 


\begin{figure}[!htp]
\centering
\includegraphics[width=.45\textwidth]{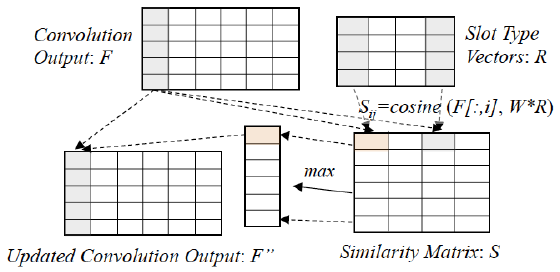}
\vspace{-0.2cm}
\caption{Global Attention.}
\label{golbalattention}
\vspace{-0.2cm}
\end{figure}

We apply local attention to each convolution output of each subgraph, 
then take the concatenation of three flattened attentive pooling outputs to a fully connected layer and generate a robust feature representation. Similarly, another feature representation is generated based on global attention. We concatenate these two features 
to the softmax layer to get the predicted types.   


 
\section{Experiments}


\subsection{Data}
\label{sec:data}


For model training, \newcite{angeli2014combining} created some high-quality clean annotations for SF based on crowd-sourcing\footnote{http://nlp.stanford.edu/software/mimlre-2014-07-17-data.tar.gz}. In addition,~\newcite{adel2016comparing} automatically created a larger size of noisy training data 
based on distant supervision, including about 1,725,891 positive training instances for 41 slot types. 
We manually assessed the correctness of candidate filler identification and their slot type annotation, and extracted a subset of their noisy annotations and combined it with the clean annotations. 
Ultimately, we obtain 23,993 positive and 3,000 negative training instances for all slot types.

We evaluate our approach in two settings: (1) relation extraction for all slot types, given the boundaries of query and candidate fillers. We use a script\footnote{http://cistern.cis.lmu.de.} to generate a test set (4892 instances) from KBP 2012/2013 slot filling evaluation data sets with manual assessment. (2) apply our approach to re-classify and validate the results of slot filling systems. 
We use the data from the KBP 2013 Slot Filling Validation (SFV) shared task, which consists of merged responses returned by 52 runs from 18 teams submitted to the Slot Filling task.




We used the May-2014 English Wikipedia dump to learn word embeddings based on the Continuous Skip-gram model~\cite{mikolov2013distributed}. Table~\ref{parameters} shows the hyper-parameters that we use to train embeddings and our model.

\begin{table}[h]
\small
\begin{tabular}{p{1.3cm}<{\centering}|p{4.5cm}<{\centering}|p{0.7cm}<{\centering}}
\hline 
Parameter & Parameter Name & Value \\
\hline
$d$ & Word Embedding Size & 50 \\
$\lambda$ & Initial Learning Rate & 0.1 \\
$\widehat{K}$ & \# of Filters in Convolution Layer & 500 \\
$h$ & Hidden Unit Size in Fully Connected Layer & 1000 \\
$k^{p}$ & Max Pooling Size & 3 \\
\hline
\end{tabular}
\vspace{-0.2cm}
\caption{Hyper-parameters.}
\label{parameters}
\end{table}
\vspace{-0.1cm}

\begin{table*}[h]
\footnotesize
\begin{tabular}{p{1.1cm}|p{2.6cm}|p{6.9cm}|p{3.8cm}}
\hline 
& Method & Description & Features \\
\hline
\multirow{3}{1.1cm}{Previous Methods} & FCM~\cite{yu2014factor} & A factor-based compositional embedding model by deriving sentence-level and substructure representations & word embedding, dependency parse, WordNet, name tagging \\
\cline{2-4}
& CR-CNN~\cite{santos2015classifying} & Applying a pairwise ranking loss function over CNNs & word embedding, word position embedding \\
\cline{2-4}
& Context-CNN~\cite{adel2016comparing} & Splitting each sentence into three parts based on query and filler positions, and apply a CNNs to each part & word embedding \\
\hline
\hline

\multirow{6}{1.1cm}{Our Methods} & DepCNN & Applying CNNs to the shortest dependency path between query and filler & word embedding, dependency parse \\
\cline{2-4}
 & GraphCNN & DepCNN + applying CNNs to both query and filler related contextual graphs & word embedding, dependency parse \\
\cline{2-4}
& GraphCNN+L & incorporating query and filler information as local attention into the GraphCNN & word embedding, dependency parse \\
\cline{2-4}
& GraphCNN+G$^{1}$ & incorporating slot type representations learned from type names as global attention into the GraphCNN & word embedding, dependency parse \\
\cline{2-4}
& GraphCNN+G$^{2}$ & incorporating slot type representations learned from external KB as global attention into the GraphCNN & word embedding, dependency parse, knowledge base \\
\cline{2-4}
& GraphCNN+L+G$^{2}$ & incorporating both local and KB based global attentions into the GraphCNN & word embedding, dependency parse, knowledge base \\
\hline
\end{tabular}
\vspace{-0.2cm}
\caption{Approach Descriptions for Multi-Class Relation Classification}
\label{baseline}
\end{table*}

\subsection{Relation Extraction}
\label{sec:comparison1}


We compare with several existing state-of-the-art slot filling and relation extraction methods. Besides, we also design several variants to demonstrate the effectiveness of each component in our approach. Table~\ref{baseline} presents the detailed approaches and the features used by these methods.


We report scores with Macro F$_1$ and Micro $F_1$. 
Macro F$_1$ is computed from the average precision and recall of all types while Micro F$_1$ is computed from the overall precision and recall, 
which is more useful when the size of each category varies. Table~\ref{classification} shows the comparison results on relation extraction. 





\begin{table}[h]
\small
\begin{tabular}{p{1.0cm}|p{2.3cm}|p{1.3cm}<{\centering}|p{1.3cm}<{\centering}} 
\hline 
& Method & Micro F1 & Macro F1 \\
\hline
\multirow{3}{1.0cm}{Previous Methods} & FCM & 41.13 & 12.68 \\
& CR-CNN & 41.61 & - \\
& ContextCNN & 41.31 & 29.01 \\
\hline
\multirow{5}{1.0cm}{Variants of Our Methods}& DepCNN & 54.91 & 36.63 \\
& GraphCNN & 55.63 & 36.74 \\
& GraphCNN+L & 56.29 & 37.12 \\
& GraphCNN+G$^{1}$ & 56.18 & 36.87 \\
& GraphCNN+G$^{2}$ & 56.81 & 38.15 \\
\hline
Our Method & GraphCNN+L+G$^{2}$ & \textbf{57.39} & \textbf{38.26} \\
\hline
\end{tabular}
\caption{Relation Extraction Component Performance on \textbf{Slot Filling Data Set} (\%).}
\label{classification}
\vspace{-0.2cm}
\end{table}

We can see that by incorporating the shortest dependency path or regularized dependency graph into neural networks, the model can achieve more than 13\% micro F-score gain over the previously widely adopted methods by state-of-the-art systems for SemEval relation classification. It confirms our claim that SF is a different and more challenging task than traditional relation classification and also demonstrates the effectiveness of dependency knowledge for SF. 

In addition, by incorporating local or global attention mechanism into the GraphCNN, the performance can be further improved, which proves the effectiveness of these two attention mechanisms. Our method finally achieves absolute 16\% F-score gain by incorporating the regularized dependency graph and two attention mechanisms. 

To better quantify the contribution of different attention mechanisms on each slot type, we further compared the performances on each single slot type. Table~\ref{comparison} shows the gain/loss percentage of the Micro F1 by adding local attention or global attention into GraphCNN for each slot type. We can see that both attentions yield improvement for most slot types. 


\begin{table*}[h]
\small
\begin{tabular}{p{2.7cm}|p{1.0cm}<{\centering}|p{1.6cm}<{\centering}||p{2.1cm}<{\centering}|p{0.8cm}<{\centering}||p{2.1cm}<{\centering}|p{2.4cm}<{\centering}}
\hline 
\multirow{2}{*}{Slot Type} & \multicolumn{2}{p{3.0cm}||}{Impact of Attention (\%)} & \multirow{2}{2.1cm}{Training Data Distribution (\%)} & \multirow{2}{0.8cm}{F1 (\%)} & \multirow{2}{2.1cm}{Wide Context Distribution (\%)} & \multirow{2}{2.4cm}{Impact of Depen- dency Graph (\%)} \\
\cline{2-3}
 & Local & Global-KB &  &  &  &  \\
\hline

state\_of\_death & 9.8 & -0.4 & 0.9 & 41.8 & 66.7 & 44.2 \\
date\_of\_birth & 7.3 & 121.3 & \textbf{1.3} & \textbf{84.1} & \textbf{20.0} & \textbf{-81.9} \\
age & 4.1 & -5.3 & \textbf{1.3} & \textbf{98.5} & 15.9 & 28.5 \\
per:alternate\_names & -2.0 & 21.2 & 1.5 & 36.6 & 41.5 & 62.0 \\
origin & -0.9 & 7.8 & 1.7 & 61.5 & 29.3 & 137.3 \\
country\_of\_birth & 16.7 & 12.0 & 1.9 & 61.5 & 55.6 & 162.5 \\
city\_of\_death & 1.1 & 3.3 & 1.9 & 61.3 & 70.3 & 24.4 \\
state\_of\_headq. & 9.7 & -5.1 & \textbf{3.1} & \textbf{51.7} & 54.8 & 95.7 \\
cities\_of\_residence & 4.5 & 5.7 & 3.5 & 57.3 & 77.0 & 40.5 \\
states\_of\_residence & -4.3 & 2.3 & 3.8 & 50.5 & \textbf{45.9} & \textbf{175.8} \\
country\_of\_headq. & 5.6 & -0.8 & \textbf{5.3} & \textbf{41.5} & 54.4 & 146.3 \\
city\_of\_headq. & 1.6 & -6.9 & \textbf{6.7} & \textbf{30.3} & 54.9 & 39.3 \\
employee\_\_of & 14.9 & 4.9 & 7.3 & 65.9 & \textbf{54.9} & \textbf{132.5} \\
countries\_of\_residence & 37.7 & 8.6 & 7.4 & 47.4 & 47.2 & 134.9\\

\hline
\end{tabular}
\caption{Comparison Analysis for Each Slot Type.}
\label{comparison}
\vspace{-0.2cm}
\end{table*}

\subsection{Slot Filling Validation}

\begin{figure}[!htp]
\centering
\includegraphics[width=.38\textwidth]{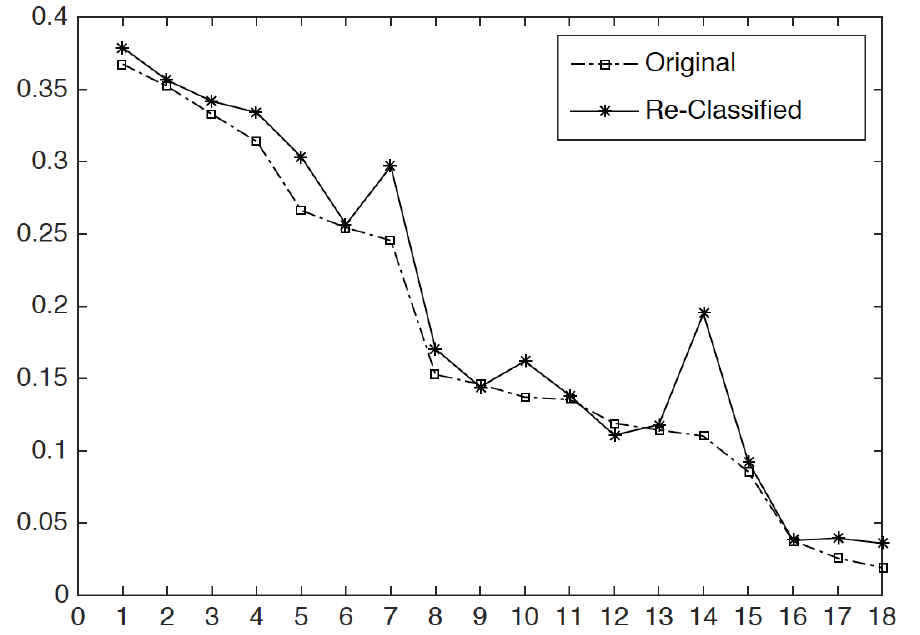}
\vspace{-0.2cm}
\caption{Comparison on Individual System}
\label{sfv}
\vspace{-0.1cm}
\end{figure}

In TAC-KBP 2013 Slot Filling Validation (SFV)~\cite{jioverview} task, there are 100 queries. 
We first retrieve the sentences from the source corpus (about 2,099,319 documents) and identify the query and candidate filler using the offsets generated by each response, then apply our approach to re-predict the slot type. 
Figure~\ref{sfv} shows the F-scores based on our approach and the original system. 
For a system which has multiple runs, we select one for comparison. We can see that our approach consistently improves the performance of almost all SF systems in an absolute gain range of [-0.18\%, 8.48\%]. 
With analysis of each system run, we find that our approach can provide more gains to the SF systems which have lower precision. 


Previous studies~\cite{tamang2011adding,rodriguez2015university,zhi2015modeling,viswanathan2015stacked,rajanicombining,yu2014wisdom,rajani2016supervised} for SFV trained supervised classifiers based on features such as confidence score of each response and system credibility. 
For comparison, we developed a new SFV approach: a new SVM classifier based on a set of features (docId, filler string, original predicted slot type and confidence score, new predicted slot type confidence score based on our neural architecture) for each response 
to take advantage of the redundant information from various system runs. Table~\ref{sfvPerformance} compares our SFV performance against previous reported scores on judging each response as true or false. We can see that our approach advances state-of-the-art methods.
\begin{table}[h]
\small
\begin{tabular}{p{3.2cm}|p{1.1cm}<{\centering}|p{0.8cm}<{\centering}|p{1.0cm}<{\centering}}
\hline 
Methods & Precision (\%) & Recall (\%) & F-score (\%) \\
 \hline
Random & 28.64  & 50.48  & 36.54 \\
Voting & 42.16  & 70.18  & 52.68 \\
Linguistic Indicators & 50.24  & 70.69  & 58.73 \\
SVM  & 56.59  & 48.72  & 52.36 \\
MTM~\cite{yu2014wisdom} &  53.94 & 72.11  & 61.72 \\
\hline
Our Approach & \textbf{70.46} & 64.07 & \textbf{67.11} \\
\hline
\end{tabular}
\caption{Overall Performance for SFV: all the Baseline Systems are from~\newcite{yu2014wisdom}.}
\label{sfvPerformance}
\end{table}


\subsection{Detailed Analysis}
\label{sec:analysis}





\vpara{Significance Test:} Table~\ref{classification} shows the results of multiple variants of our approach. To demonstrate the difference between the results of these approaches are not random, we randomly sample 10 subsets (each contains 500 instances) from the testing dataset, and conduct paired t-test between each of these two approaches over these 10 data sets to check whether the average difference in their performances is significantly different or not. Table~\ref{significance} shows the two-tailed P values. The differences are all considered to be statistically significant while all p-values are less than 0.05.



\begin{table}[h]
\small
\begin{tabular}{p{2.5cm}|p{2.5cm}|p{1.3cm}<{\centering}} 
\hline 
Method 1 & Method 2 & P Value \\
\hline
DepCNN & GraphCNN &  0.0165\\
GraphCNN & GraphCNN+L & 0.0007  \\
GraphCNN & GraphCNN+G$^{1}$ & 0.0160 \\
GraphCNN & GraphCNN+G$^{2}$ & $<$0.0001\\
GraphCNN+L & GraphCNN+L+G$^{2}$ & 0.0009 \\
GraphCNN+G$^{2}$ & GraphCNN+L+G$^{2}$ & 0.0010 \\
\hline
\end{tabular}
\caption{Statistical Significance Test.}
\label{significance}
\vspace{-0.2cm}
\end{table}

\vpara{Impact of Training Data Size:} We examine the impact of the size of training data on the performance for each slot type. Table~\ref{comparison} shows the distribution of training data and the F-score of each single type. We can see that, for some slot types, such as \textit{per:date\_of\_birth} and \textit{per:age}, the entity types of their candidate fillers are easy to learn and differentiate from other slot types, and their indicative words are usually explicit, thus our approach can get high f-score with limited training data (less than 507 instances)
. In contrast, for some slots, such as \textit{org:location\_of\_headquarters}, their clues are implicit and the entity types of candidate fillers are difficult to be inferred. Although the size of training data is larger (more than 1,433 instances), the f-score remains quite low. One possible solution is to incorporate fine-grained entity types from existing tools into the neural architecture. 

\vpara{Impact of Wide Context Distribution:} We further compared the performance and distribution of instances with wide contexts across all slot types. A context is considered as wide if the query and candidate filler are separated with more than 7 words.
The last column of Table~\ref{comparison} shows the performance by incorporating regularized dependency graph (ContextCNN \textit{v.s.} GraphCNN). We can see that, for most slot types with wide contexts, such as \textit{per:states\_of\_residence} and \textit{per:employee\_of}, the f-scores are improved significantly while for some slots such as \textit{per:date\_of\_birth}, the f-scores decrease because  
most date phrases do not exist in our pre-trained embedding model.

\vpara{Error Analysis:} Both of the relation extraction and SFV results showed that, more than 58\% classification errors are spurious.
Besides, we also observed many misclassifications that are caused by conflicting clues. There may be several indicative words within the contexts, but only one slot type is labeled, especially between \textit{per:location\_of\_death} and \textit{per:location\_of\_residence}. For example, in the following sentence: 
\begin{enumerate}[label={E}\arabic*.]\setcounter{enumi}{3}
\item \textit{\textbf{Billy Mays}$_{query}$, a beloved and parodied pitchman who became a pop-culture figure through his commercials for cleaning products like Orange Glo, OxiClean and Kaboom, died Sunday at his home in \textbf{Tampa}$_{filler}$, Fla.}, 
\vspace{-0.5cm}
\end{enumerate}
the correct slot type is \textit{per:city\_of\_death} while our approach mistakenly labeled it as \textit{per:city\_of\_residence} with clue words like \textit{home}. In addition, as we mentioned before, slot typing heavily relies on the fine-grained entity type of candidate filler, especially for the \textit{location} (including \textit{city, state, country}) related slot types. When the context is not specified enough, we can only rely on the pre-trained embeddings of candidate fillers, which may not be as informative as we hope. Such cases will benefit from introducing additional gazetteers such as \textit{Geonames}~\footnote{http://www.geonames.org/}.

\section{Related Work}
\label{sec:rel}



One major challenge of SF is the lack of labeled data to generalize a wide range of features and patterns, especially for slot types that are in the long-tail of the quite skewed distribution of slot fills~\cite{Ji2011}. Previous work has mostly focused on compensating the data needs by constructing patterns~\cite{sun2011nyu,roth2014universal}, automatic annotation by distant supervision~\cite{surdeanu2011stanfords,roth2014effective,adel2016comparing}, and constructing trigger lists for unsupervised dependency graph mining~\cite{yuunsupervised}. Some work~\cite{rodriguez2015university,zhi2015modeling,viswanathan2015stacked,rajanicombining,yu2014wisdom,rajani2016supervised,ma2015faitcrowd} also attempted to validate slot types by combining results from multiple systems.

Our work is also related to dependency path based relation extraction. The effectiveness of dependency features for relation classification has been reported in some previous work~\cite{bunescu2005shortest,zhao2005extracting,guodong2005exploring,jiang2007systematic,neville2003collective,ebrahimi2015chain,xu2015classifying,ji2014tackling}. 
\newcite{Liu2015}, ~\newcite{caibidirectional} and~\newcite{xu2015classifying} applied CNN, bidirectional recurrent CNN and LSTM to CONLL relation extraction and demonstrated that the most important information has been included within the shortest paths between entities. Considering that the indicative words may not be included by the shortest dependency path between query and candidate filler, we enrich it to a regularized dependency graph by adding more contexts.

\section{Conclusions and Future Work}

In this work, we discussed the unique challenges of slot filling compared with tradition relation extraction tasks. We  designed a regularized dependency graph based neural architecture for slot filling. By incorporating local and global attention mechanisms, this approach can better capture indicative contexts. Experiments on relation extraction and Slot Filling Validation data sets demonstrate the effectiveness of our neural architecture. In the future, we will combine additional rules, patterns, and constraints from slot filling task definition with DNN techniques to further improve slot filling.





\bibliography{emnlp2017}
\bibliographystyle{emnlp_natbib}

\end{document}